\documentclass[fleqn,10pt]{wlscirep}
\usepackage[utf8]{inputenc}
\usepackage[T1]{fontenc}
\usepackage{tikz}
\usetikzlibrary{circuits,circuits.logic}
\usetikzlibrary{circuits.logic.US}
\usepackage{amsmath,graphicx}
\usetikzlibrary{positioning}
\usepackage{subfig}

\newcounter{wavenum}

\newcommand*{\clki}{
  \draw (t_cur) -- ++(0,.3) -- ++(.5,0) -- ++(0,-.6) -- ++(.5,0) -- ++(0,.3)
    node[time] (t_cur) {};
}

\newcommand*{\bitvector}[3]{
  \draw[fill=#3] (t_cur) -- ++( .1, .3) -- ++(#2-.2,0) -- ++(.1, -.3)
                         -- ++(-.1,-.3) -- ++(.2-#2,0) -- cycle;
  \path (t_cur) -- node[anchor=mid] {#1} ++(#2,0) node[time] (t_cur) {};
}

\newcommand*{\known}[2]{
    \bitvector{#1}{#2}{white}
}

\newcommand*{\bit}[2]{
  \draw (t_cur) -- ++(0,.6*#1-.3) -- ++(#2,0) -- ++(0,.3-.6*#1)
    node[time] (t_cur) {};
}

\newcommand{\nextwave}[1]{
  \path (0,\value{wavenum}) node[left] {#1} node[time] (t_cur) {};
  \addtocounter{wavenum}{-1}
}

\newenvironment{wave}[3][clk]{
  \begin{tikzpicture}[draw=black, yscale=.7,xscale=1]
    \tikzstyle{time}=[coordinate]
    \setlength{\unitlength}{1cm}
    \def\wavewidth{#3}
    \setcounter{wavenum}{0}
    \nextwave{#1}
    \foreach \t in {0,1,...,\wavewidth}{
      \draw[dotted] (t_cur) +(0,.5) node[above] {t=\t} -- ++(0,.4-#2);
      \clki
    }
}{\end{tikzpicture}}

\title{Stochastic-based Neural Network hardware acceleration for an efficient virtual screening}

\author[1]{Christian F. Frasser}
\author[2,4]{Carola de Benito}
\author[3,4]{Vincent Canals}
\author[1,4]{Miquel Roca}
\author[5]{Pedro J. Ballester}
\author[1,4]{Josep L. Rosselló*}

\affil[1]{Grup d'Enginyeria Electrònica, Physics Dept., Universitat de les Illes Balears, Palma de Mallorca, 07122, Spain}
\affil[2]{Grup de Sistemes Electrònics, Physics Dept., Universitat de les Illes Balears, Palma de Mallorca, 07122, Spain}
\affil[3]{Grup d'Enginyeria Mecànica, Physics Dept., Universitat de les Illes Balears, Palma de Mallorca, 07122, Spain}
\affil[4]{Balearic Islands Health Research Institute, 07010 Palma de Mallorca, Spain.}
\affil[5]{Cancer Research Center of Marseille, Institut Paoli-Calmettes, Aix-Marseille Université, Marseille, F-13284, France}

\affil[*]{Corresponding author: j.rossello@uib.es}

\begin{abstract}
Artificial Neural Networks (ANN) have been popularized in many science and technological areas due to their capacity to solve many complex pattern matching problems. The use of ANN may have a considerable impact in those areas dealing with huge ammounts of data as is the case of Virtual Screening, a research area that studies how to identify those molecular compounds with the highest probability to present biological activity for a therapeutic target. Due to the vast number of small organic compounds and the thousands of targets for which such large-scale screening can potentially be carried out, there has been an increasing interest in the research community to increase both processing speed and energy efficiency in the screening of molecular databases. In this work build a classification model describing each molecule with a single energy-based 12D vector.  
Along with the energy-based model, we propose a machine-learning system based on the use of ANNs. Different ANNs with a number of synapsis ranging between 385 to up to 6657 are studied with respect to their suitability to identify biochemical similarities. Also, a hardware acceleration platform based on the use of stochastic computing is proposed for the ANN implementation. This platform may be useful if a high energy-efficiency and processing speed are needed when screening vast libraries of compounds. 
As a result, the proposed model showed appreciable improvements over other ligand-based methods present in the literature, with an AUC performance of 0.83 when applied to the DUD-E data-set, and an Enrichment Factor at the 1\% ($EF_{1\%}$) of  20.71. For the case of the hardware accelerator, it is able to achieve a processing speed over 73,000 inferences per second with an energy efficiency of 3500 inferences per Joule consumed.
For the special case of implementing an specific ANN training per therapeutic target, the AUC value increases from 0.83 to up to a 0.94 value while the Enrichment Factor ($EF_{1\%}$) increases to 30.14.
To summarize, the proposed energy-based binary classification model, in combination with the use of ANN accelerators may considerably improve the virtual screening performance in terms of accuracy, speed and energy-efficiency.

\end{abstract}
\begin{document}

\flushbottom
\maketitle
\thispagestyle{empty}

\section*{Introduction}

Artificial Neural Networks (ANN), inspired in the way in which the information is processed in the brain, have been popularized in many science and technological areas due to its capacity to solve many complex pattern matching problems. Due to its generic nature, these systems can be applied to adjust complex relationships without considering the underlying physical links since only a sufficiently large data-set of samples are needed to correlate data (represented as the input and the desired output of the ANN). 

ANNs are normally structured as a set of interconnected elementary processing units, each one implementing a non-linear operation. The non-linearity is understood as the fact that the results obtained when activating different stimulus to a system at the same time differ from the addition of the single responses that arise when each stimulus is activated separately. A simple example of the intrinsic non-linearity of real-life systems is the difference between the moves that arise when different pool players play separately with one single ball or at the same time. The possibility of interaction between simultaneous plays implies that the global behavior cannot be described from knowing how they behave separately. Mathematically this is expressed as $f(x+y) \neq f(x)+f(y)$ and the main consecuence is a much larger range of different behaviors in comparison to the case of linear relationships. For the case of a neuron with $N$ inputs inside the ANN, instead of having a total of $\alpha N$ degrees of freedom for the possible responses of the neuron (in case it was simply a linear neuron), there is exponentially more diversity. The response of a single neuron inside an ANN is normally following the expression $x_i=f \left( \sum_j \omega_{ij} I_j \right)$ , where $f\left(\cdot \right)$ is a non-linear function, $I_j$ is the $j^{th}$ input (that can come from the outside of the network or from the output of any other neuron of the network) and parameter $\omega_{ij}$ is a convenient weight dependent on the desired functionality and that must be properly adjusted from the available data. Due to the huge configurability of an ANN, a large number of training examples are needed for the adjustment of the connectivity matrix $\Omega=\left\{\omega_{ij}\right\}$, that is specially easy to obtain in case a Feed-Forward-Neural-Network (FFNN) topology is used (characterized to not have feed-back inside the network). Given an specific training data-set that correlate the desired network response with respect to the inputs, the optimization of $\Omega$ is a non-convex problem that may be solved using back-propagation. 

One of the key-points of creating accurate ANNs is the availability of a sufficiently large amount of data for its proper training. Given the data explosion that has been present at all the science and technology areas (and chemistry is not an exception), the use of ANNs arises as an optimum way to convert data to useful information. Moreover, data explosion is specially critical in organic chemistry due to the truly vast possibilities for constructing chemical compounds \cite{Hoffmann20191148}. This huge phase space can be seen in the ever increasing number of compounds included in different chemical databases, that can now be of the order of billions. For the specific problem of the analysis of interactions between different compounds, that is the aim of any drug discovery process, the number of possibilities explode and can be unmanageable in the case that we want to explore all the cases. For these huge problems, the direct implementation in hardware of high-performance and energy-efficient ANNs may be a solution.

Neuromorphic hardware (NH) is an increasing research field pushed by the need of developing high-performance Artificial Intelligence (AI) systems, with a potential capacity to provide timely responses to those applications requiring to process huge amounts of data \cite{Morro20181371}. Many efforts have been made in NH using digital \cite{Nascimento2013} or analogue \cite{Carrasco-Robles2009} circuits. Nevertheless, the inclusion of large amounts of multipliers (to reproduce the neural weighting) unable the proper parallelization of complex networks (and therefore to speed-up the process). A possible feasible solution for this is the use of approximate multipliers, that may be built using Stochastic Computing (SC) concepts \cite{Gaines1967149,Alaghi2013}. The representation of data in the SC paradigm is performed in a probabilistic way with the use of boolean quantities that are aleatory switching during time. The probability of a single bit to be in a given state is used to encode the desired signal, therefore one single bit is enough to carry this probabilistic information within the hardware instead of the set of bits (bus), since information is codified through time. With this probabilistic codification, complex functions normally implemented with large amounts of resources (as the case of binary multipliers) may be performed by using a single logic gate, with great savings in terms of area and power dissipation. However, this area reduction has a cost associated in terms of loss of precision, that may be not critical in most machine learning applications where a relatively reduced set of output possibilities (the categories) must be distinguished based on generic similarities. In fact, most of the current machine-learning applications use a low number of bits to represent digitized signals since the difference in the final result between using high-precision floating-point signals or single precision 8/16 bits signals is negligible. Therefore, when implementing low-precision calculations in SC, the integration time used to evaluate the result of stochastic operations may be considerably reduced. At the same time, and due to the low gate count needed in SC operations, a high-parallelism may be implemented in a single chip, that is of special interest in ANNs due to its intrinsic parallel nature. Therefore, we can understand Stochastic Computing as a natural way to implement ANNs and to efficiently exploit its amazing capacities.
 
In this work we present an stochastic-based neuromorphic hardware to accelerate ANNs for Virtual Screening. The primary goal of Virtual Screening (VS) \cite{Schneider2010273,Singh2020,Glaab2016352} is to retrieve a small subset of molecules with the highest possible proportion of actives in the screened library. When a 3D structure of the target is available and the binding site is known, this problem is more specifically called structure-based VS\cite{Lih2020,Batool2019,Pinzi2019}. On the other hand, with at least a molecule with activity for the target, methods for ligand-based VS can be used\cite{Zoete20161399,Li2016W436,Kumar2018}. A particular class of ligand-based methods that can be used for virtual screening exploits, instead of a single ligand used as a search template, a set of molecules\cite{Neves2018,Soufan2018,Speck-Planche2012273}. Such models are typically generated with machine learning trained on ligands with known activity for the target and their chemical properties\cite{Olier2018285,Sidorov2020}.

Here we use a set of energy-based molecular descriptors\cite{Oliver2017} are used as inputs to the ANN system and tested by analyzing the DUD-E data-set of chemical compounds. \cite{Mysinger20126582} A predictive performance similar or higher than other ligand-based models is obtained, along with a good performance in terms of both processing speed and energy-efficiency. 

\section*{Methods}

\subsection*{Compound description through Molecular Pairing Energies}
In this work we propose a classification model in which we assume a relationship between chemical structures described by a set of physico-chemical parameters (what we call the molecular pairing energies MPE) and the biological activity. The model will be based on the use of a multi-layer neural network to predict the activities of chemicals from its MPE values.  
The compounds’ pairing energies are defined for every pair of atoms of the molecule by using the partial charges $q_i$ and $q_j$ of each atom and the distance between them $r_{ij}$ as follows:

\begin{equation}
\centering
    E_{ij}=K\frac{q_iq_j}{r_{ij}}
    \label{pairing}
\end{equation}

\begin{figure}
    \centering
    \includegraphics[scale=0.7]{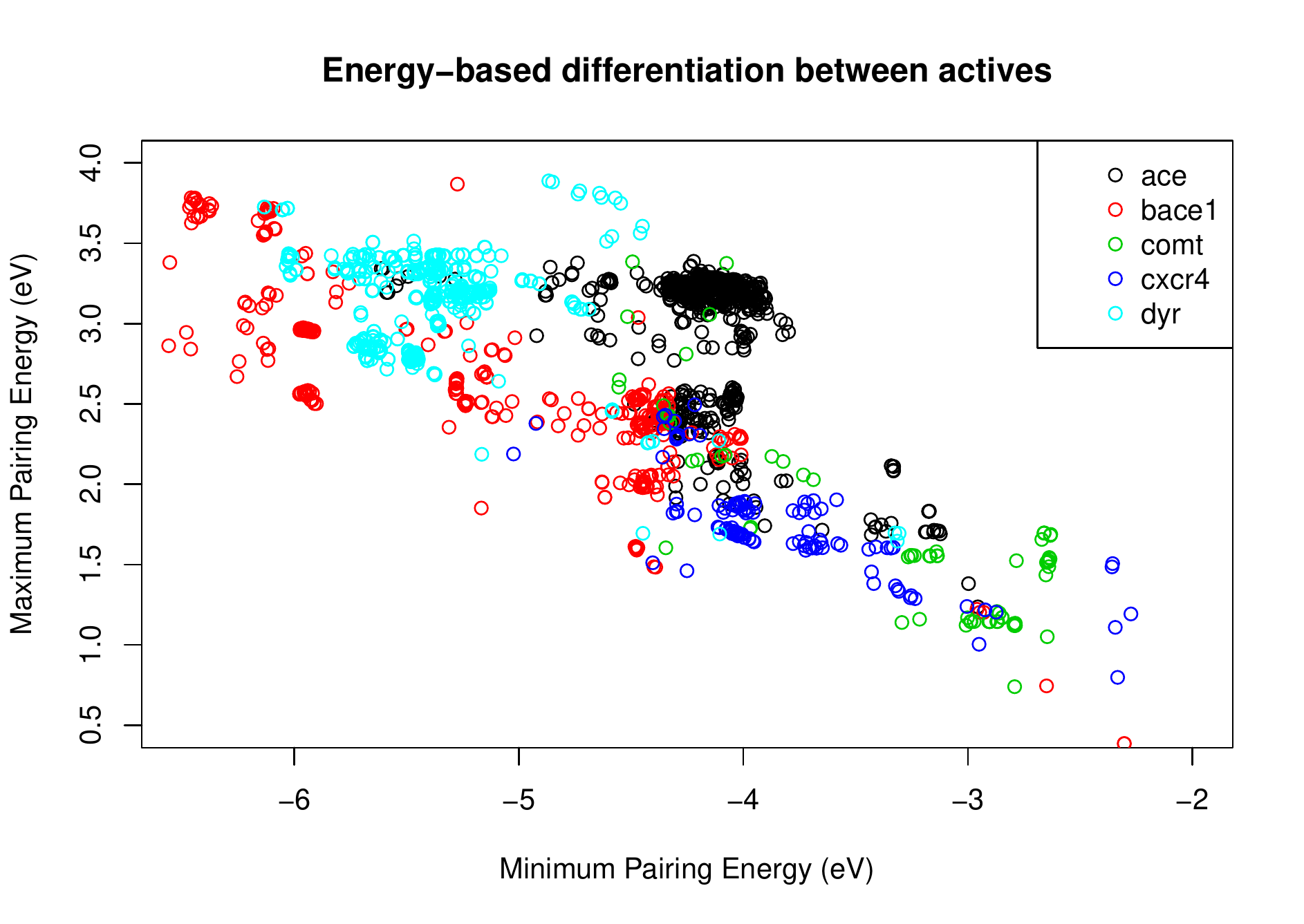}
    \caption{Clustering capacity of pairing energy descriptors when using the most negative and most positive MPE values for five different DUD-E actives.}
    \label{fig:Clustering}
\end{figure}

From all the pairing energies present in a compound, we chose the six highest positive  and the six lowest negative energies, thus creating a 12D vector for the description of the molecule. For those compounds with less than 12 possible MPE values the remaining values are filled with zeros. The most electropositive or electronegative MPE values may be understood as those internal interactions more related to molecular scatter or assemble of the compound respectively. In this work, the MPE model is applied to the full DUD-E Database, in which the partial charges has been estimated by using the MMFF94 force field \cite{Halgren1996490} that is implemented within the Openbabel software. The DUD-E docking benchmark is a widely used database to evaluate virtual screening methods that is composed by 102 different pharmaceutical targets that include a crystal ligand along with a set of active compounds and decoys (assumed to not be actives). The MPE model has empirically shown a good capacity for clustering those compounds showing similar chemical properties \cite{Oliver2017}, as can be appreciated in Fig.\ref{fig:Clustering} when plotting the most positive and most negative pairing energy for five different DUD-E actives. As can be appreciated those compounds showing a higher cohesion energy also may lead to a higher scattering energy. The basic working hypothesis here is that MPE values may be efficiently applied to compare biological activities between compounds, so an AI-assisted MPE ligand-based method for Virtual Screening is developed.

\subsection*{Stochastic Computing}
Stochastic computing (SC) is characterized to represent signals using random bit-streams\cite{Gaines1967149}
. Each SC signal may be encoded by providing a value of $-1$ to zeros and $+1$ to ones, so that an $N-bit$ sequence is representing a quantity equal to $p=(N_1-N_0)/(N_0+N_1)$, being $N_0$ and $N_1$ the number of zeros and ones respectively. 
In this way, one single bit may be representing an analog quantity between $-1$ and $+1$. 
This probabilistic encoding is called \textit{bipolar} stochastic coding, where both positive and negative values can be represented.
Any complement-two coded binary number $X$ can be converted into a stochastic bit-stream $x(t)$ by comparing it with a Random Number Generator (RNG) $R(t)$. 
Therefore, each stochastic signal may be understood as a sequence of booleans $x(t)=\{X>R(t)\}$, and the precision of these signals is related to the total time length considered (evaluation time). 
The generation of the aleatory variable $R(t)$ must be uniform in the interval of all possible values of $X$ (from $-2^{(b-1)}$ to $2^{(b-1)}-1$ for a binary signal with a total of $b$ bits in 2's-complement codification). 
In order to recover the original value $X$ in the binary domain, a signed up/down counter and a register may be used.

\begin{figure}
\centering
\begin{tikzpicture}[circuit logic US,scale=1.1,line width=1.5pt]
\node (Ai) at (0,0){$X$};
%\draw[fill=black!5,very thick,dashed] ($(Ai)+(0.2,0.7)$) -- ($(Ai)+(1.8,0.7)$)--($(Ai)+(1.8,-1.5)$) -- ($(Ai)+(-0.6,-1.5)$)  -- ($(Ai)+(-0.6,-0.5)$) -- ($(Ai)+(0.2,-0.5)$) --cycle;
\node (Ri) at ($(Ai)+(-0.2,-1)$) {$R(t)$};
\draw[ultra thick] ($(Ai)+(0.2,0)$) -- ($(Ai)+(0.8,0)$);
\draw[ultra thick] ($(Ai)+(0.2,-1)$) -- ($(Ai)+(0.8,-1)$);
\draw[fill=white!20,ultra thick] ($(Ai)+(0.5,0.5)$) -- ($(Ai)+(1.5,0)$) -- ($(Ai)+(1.5,-1)$)  -- ($(Ai)+(0.5,-1.5)$) -- cycle;
\node (sml) at ($(Ai)+(1,-0.5)$) {$>$};
\node [xnor gate,fill=white!40,thin] (a1) at ($(Ai)+(3,-0.6)$) {};
\node (a) at ($(a1)+(-1,0.11)$){};
\node (b) at ($(a1)+(-1,-0.11)$){}; 
\node (as) at ($(a)+(0.2,0.3)$){\large$x(t)$};
\node (bs) at ($(b)+(0.2,-0.3)$){\large$y(t)$}; %\node (out) at ($(Ai)+(1,-3.11)+(1,0.3)$){\large$z(t)$}; 
\node (out) at ($(a1.output)+(0.3,0.3)$){\large$z(t)$}; 
\draw[thin] ($(sml)+(0.5,0)$) -- ++(right:3mm)  |- ($(a1.input 1)$);
\draw[color=black,thin] (b) -- ++(right:3mm) --(a1.input 2);
\draw[color=black,thin] (a1.output) -- ++(right:3mm) ;
\draw[fill=white!20,ultra thick] ($(a1.output)+(1.5,1.5)+(-0.3,-1)$) -- ($(a1.output)+(1.5,1.5)+(0.7,-1)$) -- ($(a1.output)+(1.5,1.5)+(0.7,-2.5)$)  -- ($(a1.output)+(1.5,1.5)+(-0.3,-2.5)$) -- cycle;
\node (bs2) at ($(a1.output)+(1.5,1.5)+(1,0)$) {};
\draw[fill=white!20,ultra thick] ($(bs2)+(0.7,-1)$) -- ($(bs2)+(1.7,-1)$) -- ($(bs2)+(1.7,-2.5)$)  -- ($(bs2)+(0.7,-2.5)$) -- cycle;
\node (sml2) at ($(a1.output)+(1.5,0.1)$) {$u/d$};
\node (cnt) at ($(sml2)+(0.3,-0.3)$) {$count$};
\node (clk) at ($(sml2)+(-0.2,-0.6)$) {$>$};
\draw[thin] ($(clk)+(-0.5,0)$) -- ($(clk)+(-0.1,0)$) ;
\node (clksmbl) at ($(clk)+(-0.7,0)$) {$clk$};
\draw[ultra thick] ($(cnt)+(0.4,0)$) -| ($(cnt)+(1,0.4)$) -- ($(cnt)+(1.4,0.4)$) ;
\node (Z) at ($(cnt)+(0.7,0.3)$) {$Z$};
\node (D) at ($(cnt)+(1.6,0.4)$) {$D$};
\node (Q) at ($(D)+(0.6,0)$) {$Q$};
\node (clk2) at ($(sml2)+(1.8,-0.6)$) {$>$};
\draw[thin] ($(a1.output)+(0.3,0)$) |- ($(sml2)+(-0.3,-0.1)$);
\end{tikzpicture}
\begin{wave}{9}{7}
\nextwave{$Enable_{FFD}$} \bit{0}{7} \bit{1}{1}
\nextwave{X} \known{0.000 (binary, 2's-complement)}{8}
\nextwave{R(t)} \known{0.010}{1}\known{1.010}{1}\known{1.110}{1}\known{0.110}{1}\known{1.001}{1}\known{0.111}{1}\known{1.000}{1}\known{0.011}{1}
\nextwave{$x=0.0$} \bit{0}{1} \bit{1}{2} \bit{0}{1} \bit{1}{1} \bit{0}{1} \bit{1}{1} \bit{0}{1}
\nextwave{$y=0.5$} \bit{1}{1} \bit{0}{1} \bit{1}{3} \bit{0}{1} \bit{1}{2}
\nextwave{$z=0.0$} \bit{0}{2} \bit{1}{1} \bit{0}{1} \bit{1}{3} \bit{0}{1} 
\nextwave{Z} \known{1.111}{1}\known{1.110}{1}\known{1.111}{1}\known{1.110}{1}\known{1.111}{1}\known{0.000}{1}\known{0.001}{1}\known{0.000}{1}
\nextwave{Q} \known{X}{8}\known{0.000}{1}
\end{wave}
\caption{Example of an SC \textit{bipolar} multiplier. Two switching inputs $x$ and $y$ are representing signals $0$ and $0.5$ respectively and multiplied through the XNOR gate, leading to signal $z=0.0$. The conversion of signal $X$ from the binary to the stochastic domain is performed using a RNG $R(t)$ and a comparator. 
The conversion of an stochastic number $z$ to the binary domain is performed using a signed up/down counter and a register. The register is enabled only at the end of the evaluation time. 
For a proper operation, signals $x(t)$ and $y(t)$ must be statistically uncorrelated. }
\label{fig.1}
\end{figure}
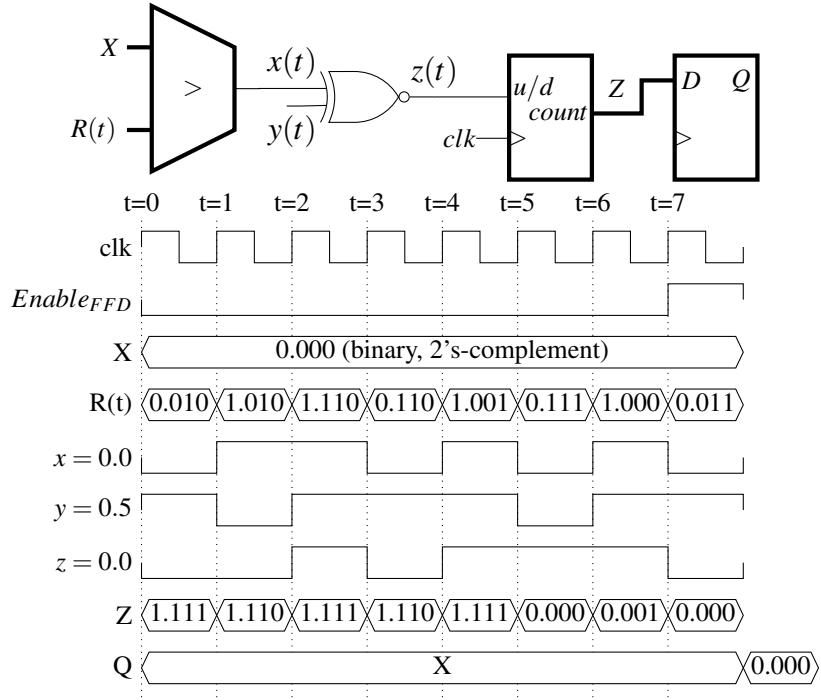

One of the advantages of SC is the low cost in terms of hardware resources to implement complex functions. The most clear example is the \textit{bipolar} stochastic multiplier, where the operation is computed by a single XNOR gate. In Fig.\ref{fig.1} we illustrate a multiplier to show the main parts of an SC system: the binary to stochastic conversion, the stochastic operation and a second conversion to recover a binary result. A binary signal $X$ is converted to stochastic $x(t)$ by using a time-varying random number $R(t)$ and a binary comparator.
As a result, a single stochastic bit is generated during each clock cycle, where the SC signals are tied to '1' or '0' with a given probability (note that in the example of Fig.\ref{fig.1} we show two SC signals $x(t)$ and $y(t)$ that are multiplied). 
For the stochastic product operation shown, decorrelation between signals is mandatory, that means, the covariance between signals must be zero : $Cov\big( x(t),y(t)\big) =0$. 
The output signal $z(t)$ is generated as a result of the application of the truth table of an XNOR gate (as shown in Fig.\ref{fig.1}). The final averaged value through time $Z$ is the result of the product between $x$ and $y$. 
To store the result in a memory, a conversion must be done from the time-dependent stochastic domain to the binary world. 
The conversion is easily done employing a signed up/down counter and a register to store the final value, where each $N$ clock cycles (defining the evaluation time) the register enable is set to store the expected result.

%--------------------------------------------------------------
% Figura : multiplicacion estocastica bipolar y 
%           conversion
%--------------------------------------------------------------

%--------------------------------------------------------------
% Figura : Correlacion
%           
%--------------------------------------------------------------
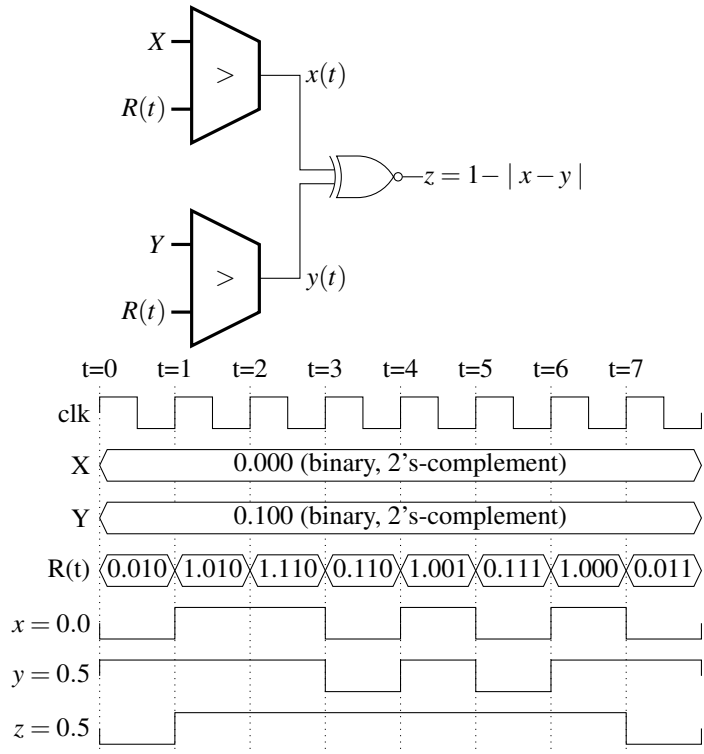
\begin{figure}
\centering
\begin{tikzpicture}[circuit logic US,scale=0.9]
\node (Ai) at (0,0){$X$};
\node (Ri) at ($(Ai)+(-0.2,-1)$) {$R(t)$};
\draw[very thick] ($(Ai)+(0.2,0)$) -- ($(Ai)+(0.8,0)$);
\draw[very thick] ($(Ai)+(0.2,-1)$) -- ($(Ai)+(0.8,-1)$);
\draw[fill=white!20,very thick] ($(Ai)+(0.5,0.5)$) -- ($(Ai)+(1.5,0)$) -- ($(Ai)+(1.5,-1)$)  -- ($(Ai)+(0.5,-1.5)$) -- cycle;
\node (sml) at ($(Ai)+(1,-0.5)$) {$>$};
\node (asc) at ($(Ai)+(2.5,-0.5)$) {$x(t)$};
\node (Ai2) at ($(Ai)+(0,-3)$){$Y$};
\draw[very thick] ($(Ai2)+(0.2,0)$) -- ($(Ai2)+(0.8,0)$);
\node (Ri2) at ($(Ai)+(-0.2,-4)$) {$R(t)$};
\draw[very thick] ($(Ai2)+(0.2,-1)$) -- ($(Ai2)+(0.8,-1)$);
\draw[fill=white!20,very thick] ($(Ai2)+(0.5,0.5)$) -- ($(Ai2)+(1.5,0)$) -- ($(Ai2)+(1.5,-1)$)  -- ($(Ai2)+(0.5,-1.5)$) -- cycle;
\node (sml2) at ($(Ai2)+(1,-0.5)$) {$>$};
\node (bsc) at ($(Ai2)+(2.5,-0.5)$) {$y(t)$};

\node [xnor gate,thin,fill=white!40] (and) at ($(Ai)+(3,-2)$) {} ;

\draw[color=black] ($(Ai)+(1.5,-0.5)$) -- ++(right:6mm) |- (and.input 1);

\draw[color=black] ($(Ai2)+(1.5,-0.5)$) -- ++(right:6mm) |- (and.input 2);

\draw[color=black] (and.output) -- ++(right:3mm) ;
\node (salida) at ($(and.output)+(1.5,0)$) {$z=1-\mid x-y \mid$};
\node (offset) at ($(and.output)+(0,-1)$) { };
\end{tikzpicture}

\begin{wave}{7}{7}
\nextwave{X} \known{0.000 (binary, 2's-complement)}{8}
\nextwave{Y} \known{0.100 (binary, 2's-complement)}{8}
\nextwave{R(t)} \known{0.010}{1}\known{1.010}{1}\known{1.110}{1}\known{0.110}{1}\known{1.001}{1}\known{0.111}{1}\known{1.000}{1}\known{0.011}{1}
\nextwave{$x=0.0$} \bit{0}{1} \bit{1}{2} \bit{0}{1} \bit{1}{1} \bit{0}{1} \bit{1}{1} \bit{0}{1}
\nextwave{$y=0.5$} \bit{1}{3} \bit{0}{1} \bit{1}{1} \bit{0}{1} \bit{1}{2}
\nextwave{$z=0.5$} \bit{0}{1} \bit{1}{6} \bit{0}{1} 
\end{wave}
\caption{Correlation between signals may change the function implemented by the logic gate. Stochastic signals $x(t)$ and $y(t)$ are said to be perfectly correlated when they share the same RNG $R(t)$. In case of correlation, an XNOR gate performs an operation related to the absolute value of the difference between signals instead of performing the product.}
\label{fig.2}
\end{figure}
%-------------------------------------------------------------------------

As has been shown in Fig.\ref{fig.1}, decorrelation between stochastic signals is necessary for some operations, as is the case of the multiplication. 
However, in presence of exact correlation (when SC signals are generated using the same RNG output), XNOR gate is no longer estimating the product operation but a function related to the absolute value of the difference between the SC signals (see Fig.\ref{fig.2}).
For quantifying the correlation, we can use the stochastic computing correlation metric, defined as:
\begin{equation}
\centering
    C(x,y)=\frac{Cov\bigl( x(t),y(t)\bigr)}{1-\mid x   -y\mid-xy}
    \label{correlacion}
\end{equation}

where function $Cov$ is the covariance between the two time-dependent stochastic signals $x(t)$ and $y(t)$ using \textit{bipolar} coding, while parameters $x$, $y$ are their averaged values (bounded between -1 and +1). 
The case $C(x,y)=+1$ implies maximum correlation (when both signals are generated from the same random number), whereas $C(x,y)=0$ implies a complete decorrelation. 
%Also, if both signals are oscillating in contra-phase, we have a correlation factor of $C(x,y)<1$. 

The output of any combinational gate can be expressed as a function of the correlation between its two inputs and also to their activities. For the case of the AND, OR and XNOR gates we have:
\begin{equation}
\centering
\begin{array}{rl}
    AND(x,y)= & \bigl(xy+x+y-1\bigr)\bigl(1-C\left( x,y\right)\bigr)\cdot0.5 +C\left( x,y\right)min\left(x,y\right)\\
    OR(x,y)= & \bigl(x+y+1-xy\bigr)\bigl(1-C\left( x,y\right)\bigr)\cdot0.5 +C\left( x,y\right)max\left(x,y\right)\\
    XNOR(x,y)= & xy\bigl(1-C\left( x,y\right)\bigr) +C\left( x,y\right)\cdot\bigl(1-\mid x-y\mid\bigr)\end{array}
\label{eq:AND-OR-XNOR}
\end{equation}

That is, when $C(x,y)=1$, an AND gate circuit performs the \textit{min} operation,  and the OR gate performs the \textit{max} function. 
For the case of an XNOR gate, the function changes from $z=x\cdot y$, when $C(x,y)=0$ to $z=1-\mid x-y\mid$ when $C(x,y)=1$. Hence, depending on the correlation between SC signals the functionality can drastically change. 

%=============================================================================
%
% NEURAL NETWORK IMPLEMENTATION
%
%=============================================================================
\subsection*{Neural Network implementation}

The main purpose of this paper is to create an accurate and energy-efficient methodology to implement a ligand-based Virtual Screening process. 
Starting from 24 MPE values, 12 per each compound to compare, we studied different FFNN models providing a single output indicating the target similarity.
Each neuron in the network computes a transfer function of its weighted inputs, so that for the $i^{th}$ neuron, the output activation is:

\begin{equation}
a_i=\phi\bigl(\sum_{j}\omega_{ij}x_j\bigr),
\label{neuron_eq}    
\end{equation}
\color{black}

where $\omega_{ij}$ is the weight assigned to the $j^{th}$ input $x_j$, with all inputs coming from the previous neural layer, and $\phi$ is the non-linear transfer function computed.
Different transfer functions have been used in literature, but we will focus on two of them: the hyperbolic tangent function ($\phi(x)=tanh(x)$) and the ReLU function ($\phi(x)=max(0,x)$).
For NN implementations, $tanh$ produces better results than ReLU function at the cost of computation efficiency. 
In this work, we exploit the benefits of these two transfer functions depending on the desired computing platform to implement (hardware ANN with high energy-efficiency or a more exact software-based ANN). In particular we use the $tanh$ for the models computed in software, focusing on the precision outcomes; whereas the ReLU function is employed for the hardware implementation due to its considerable simplicity in terms of logic gates if compared with a more complex activation function.

In Fig.\ref{fig:mlp_desc} we show the scheme of the employed ANN architecture, in which parameters $u_j^k$ refers to the $j^{th}$ component of the $k^{th}$ compound (where $k\in\{1,2\} $), $H_{l,j}$ refers to the $j^{th}$ neuron in the $l^{th}$ hidden layer, and the FFNN output $y_{out}$ is the prediction of the model. 

Stochastic computing is presented as a feasible solution to implement hardware models due to its area saving advantage when implementing operations such as the multiplication and $max$ function. In the next subsection we will explain how SC may be efficiently used to implement ANN and how to apply them for virtual screening.

%----------------------------------------------------
% FIGURA 3 : NN
%----------------------------------------------------
\tikzset{%
  every neuron/.style={
    circle,
    draw,
    minimum size=0.5cm,
  },
  every neuron_input/.style={
    circle,
    draw,
    minimum size=1cm,
    fill=blue!20
  },
  neuron missing/.style={
    draw=none, 
    scale=2,
    text height=0.333cm,
    execute at begin node=\color{black}$\vdots$
  },
}
\begin{figure}
\centering
\begin{tikzpicture}[x=1.5cm, y=1.5cm, >=stealth]

\foreach \m/\l [count=\y] in {1,2,3,missing,4,5,6,7,missing,8}
  \node [every neuron/.try, neuron \m/.try] (input-\m) at (0,2.5-\y*0.5) {};

\foreach \m [count=\y] in {1,missing,2}
  \node [every neuron/.try, neuron \m/.try ] (hidden-\m) at (2,3.3-\y*1.75) {};

\foreach \m [count=\y] in {1,missing,2}
  \node [every neuron/.try, neuron \m/.try ] (hidden2-\m) at (4,2.1-\y*1.15) {};

\foreach \m [count=\y] in {1}
  \node [every neuron/.try, neuron \m/.try ] (output-\m) at (6,0.8-\y) {};

\foreach \l [count=\i] in {1,2,3,12}
  \draw [<-] (input-\i) -- ++(-1,0)
    node [above, midway] {$u^1_{\l}$};
\foreach \l [count=\i] in {5,6,7}
  \draw [<-] (input-\l) -- ++(-1,0)
    node [above, midway] {$u^2_{\i}$};

\draw [<-] (input-8) -- ++(-1,0)
    node [above, midway] {$u^2_{12}$};

\foreach \l [count=\i] in {1,B}
  \node [above] at (hidden-\i.north) {$H_{1,\l}$};

\foreach \l [count=\i] in {1,C}
  \node [above] at (hidden2-\i.north) {$H_{2,\l}$};

\foreach \l [count=\i] in {out}
  \draw [->] (output-\i) -- ++(1,0)
    node [above, midway] {$y_{\l}$};

\foreach \i in {1,...,8}
  \foreach \j in {1,...,2}
    \draw [->] (input-\i) -- (hidden-\j);

\foreach \i in {1,...,2}
  \foreach \j in {1,...,2}
    \draw [->] (hidden-\i) -- (hidden2-\j);

\foreach \i in {1,...,2}
  \foreach \j in {1,...,1}
    \draw [->] (hidden2-\i) -- (output-\j);

\foreach \l [count=\x from 0] in {Input, Hidden, Hidden, Ouput}
  \node [align=center, above] at (\x*2,3) {\l \\ layer};

\end{tikzpicture}

\caption{Employed NN architecture for the estimation of similarity between two compounds.
Each compound $u_j^k$ is described by 12 energy descriptors, where $j$ is the $j^{th}$ component of the $k^{th}$ compound, $H_{l,j}$ refers to the $j^{th}$ neuron in the $l^{th}$ hidden layer; $B$ and $C$ are the number of neurons in the first and the second hidden layer, and the output $y_{out}$ is the prediction of the model.}
\label{fig:mlp_desc}
\end{figure}
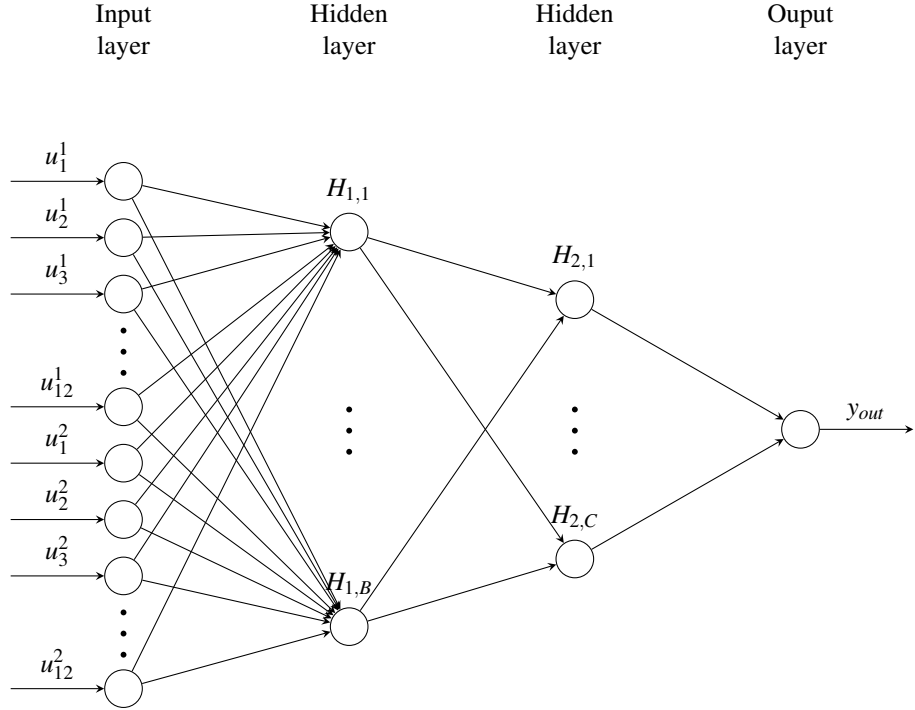

%=============================================================================
%
% Stochastic Hardware model
%
%=============================================================================
\subsection*{Stochastic Hardware model}

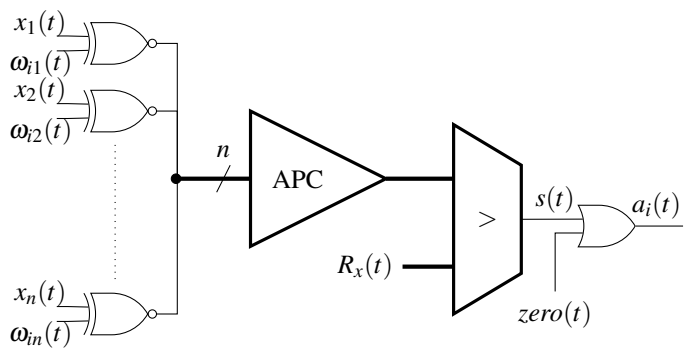
\begin{figure}
\centering
\begin{tikzpicture}[circuit logic US,scale=0.9]
%
% Portes entrades
%
\node [xnor gate,fill=white!40] (a1) at (0,0) {} ;
\node [xnor gate,fill=white!40] (a2) at (0,-1) {} ;
\draw[dotted] (0,-1.5) -- (0,-3.5) ;
\node [xnor gate,fill=white!40] (an) at (0,-4) {} ;
\draw[color=black] (a1.output) -- ++(right:3mm) |- (2,-2);
\draw[color=black] (a2.output) -- ++(right:3mm) |- (2,-2);
\draw[color=black] (an.output) -- ++(right:3mm) |- (2,-2);
\draw[color=black,ultra thick](0.9,-2) -- (2,-2);
\filldraw [black] (0.9,-2) circle [radius=2pt];
\node (bsz) at (1.6,-1.6){$n$}; 
\draw[color=black](1.5,-2.2) -- (1.7,-1.8);
%
%Labels Entrades neurona
%
\node (a) at (-1,0.11){};
\node (b) at (-1,-0.11){}; 
\node (as) at (-1.1,0.3){$x_{1}(t)$};
\node (bs) at (-1.1,-0.3){$\omega_{i1}(t)$}; 
\draw[color=black] (a) -- ++(right:3mm) -- (a1.input 1);
\draw[color=black] (b) -- ++(right:3mm) --(a1.input 2);
\node (ax) at (-1,-0.89){};
\node (bx) at (-1,-1.11){}; 
\node (asx) at (-1.1,-0.7){$x_{2}(t)$};
\node (bsx) at (-1.1,-1.3){$\omega_{i2}(t)$}; 
\draw[color=black] (ax) -- ++(right:3mm) -- (a2.input 1);
\draw[color=black] (bx) -- ++(right:3mm) --(a2.input 2);
\node (az) at (-1,-3.89){};
\node (bz) at (-1,-4.11){}; 
\node (asz) at (-1.1,-3.7){$x_{n}(t)$};
\node (bsz) at (-1.1,-4.3){$\omega_{in}(t)$}; 
\draw[color=black] (az) -- ++(right:3mm) -- (an.input 1);
\draw[color=black] (bz) -- ++(right:3mm) --(an.input 2);
%
% APC
%
\draw[fill=white!20,very thick] (2,-3) -- (2,-1) -- (4,-2) -- cycle;
\node (apc) at (2.7,-2) {APC};
%\node (log2n) at ($(apc)+(1.2,1.0)$) {$\lceil log_2(Nn)+1 \rceil$};
%\draw[->]($(log2n)+(0,-0.2)$) -- ($(log2n)+(0.55,-0.8)$);
\draw[ultra thick](4,-2) -- (5,-2);
%\draw(4.4,-2.2) -- (4.6,-1.8);

% mux
\node  at (5.5,-2.6) (A) {$>$};
\draw[very thick] ($(A)+(-0.5,-1.4)$) -- ($(A)+(-0.5,1.4)$)--($(A)+(0.5,0.8)$) -- ($(A)+(0.5,-0.8)$) -- cycle;
\node at ($(A)+(-1.8,-0.7)$) (Rt) {$R_x(t)$};
\node at ($(A)+(1.0,0.3)$) (p) {$s(t)$};
\draw [ultra thick](Rt) -- ($(Rt)+(1.3,0)$);

\node [or gate,fill=white!40] (a2) at (7.2,-2.7) {} ;
%\draw[color=black](6.45,-2.4) -- (6.7,-2.4);
%\node (rinp) at (6.2,-2.4){$0^*$};
\draw[color=black]($(a2.output)$) -- ($(a2.output)+(0.7,0)$);
\node (rinp) at (6.5,-4){$zero(t)$};
\draw[color=black] (rinp) |- (a2.input 2);

%\draw[fill=white!20,very thick] (7,-3) -- (7,-1.5) -- (7.7,-2)  -- (7.7,-2.5) -- cycle;
%\draw[ultra thick](6.5,-2.5) -- (7,-2.5);
%\node (lfsr) at (5.5,-2.8){$R(t)$}; 
%\node (lfsr) at (7.3,-2.25){$>$}; 
\draw[color=black] ($(A)+(0.5,0)$) -- (a2.input 1);
\node (sortida) at (8,-2.4){$a_{i}(t)$}; 
\end{tikzpicture}
\caption{Stochastic neuron design exploiting correlation to reduce area cost. Stochastic vectors $\mathbf{x}(t)$ and $\mathbf{w}_{i}(t)$ are uncorrelated, thus producing the stochastic multiplication with XNOR gates. An APC is used as an stochastic adder that generate a binary output consisting on the scalar poduct between vectors $\mathbf{x}(t)$ and $\mathbf{w}_i(t)$. Stochastic signals $s(t)$ and $zero(t)$ are correlated since both are generated using the same random signal $R_x(t)$, returning the \textit{max} function when evaluated through the OR gate (signal $a_i(t)$).
\label{fig_neuron_sc}}
\end{figure}

% mostrar la neurona definitiva que hemos utilizado.
The circuit implementation for each stochastic neuron is shown in Fig.\ref{fig_neuron_sc}, where the input vector $\mathbf{x}(t)$ is multiplied by the weight vector $\mathbf{w}_i(t)$ using an array of XNOR gates. The addition of these values is carried out by an Accumulative Parallel Counter (APC), which estimates the number of high values less the number of low values from the $n$ inputs and during the $N$ cycles of the whole bit-stream. 
The APC circuit produces a 2's-complement binary output representing the stochastic scalar product between $\mathbf{x}(t)$ and $\mathbf{w}_i(t)$, which must be converted again to the stochastic domain to operate in the following layers. A binary to stochastic converter is used to obtain the signal $s(t)$ (see Fig.\ref{fig_neuron_sc}). 

The ReLU transfer function is accomplished by exploiting the correlation between signals. Following the rules expressed in (\ref{eq:AND-OR-XNOR}), the $max$ function can be implemented with a single OR gate in the stochastic domain if the input signals are totally correlated $C(s,0)=1$; therefore, we use the same RNG block (generating $R_x(t)$) to convert to stochastic the APC output and to generate the stochastic zero signal $zero(t)$.

%---------------------------------------------
% FIGURA NEURONA ESTOCASTICA
%---------------------------------------------

\begin{figure}
    \centering
    \includegraphics[scale=1.0]{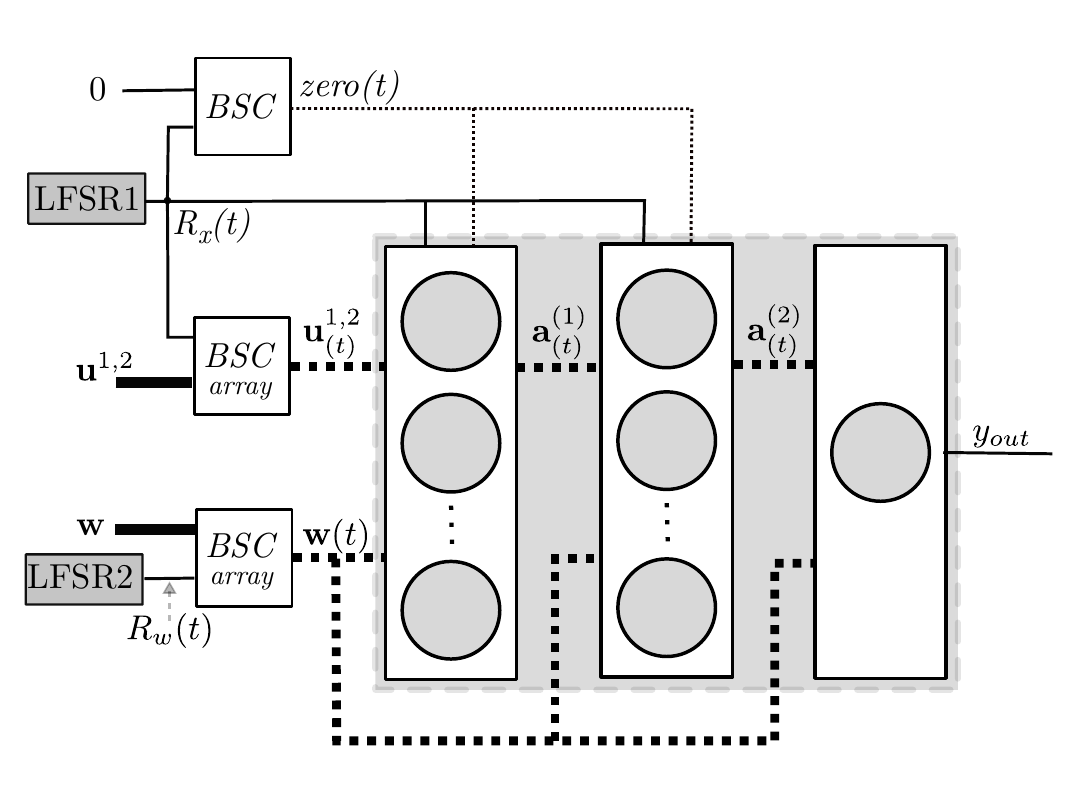}
    \caption{Neural Network implementation using two LFSR for the whole system. LFSR1 is used to generate the input stochastic vector $\mathbf{u}_{(t)}^{1,2}$, the zero \textit{bipolar} signal $zero(t)$ and the APC outcome inside each neuron. LFSR2 is used to generate the weight stochastic vector $\mathbf{w}(t)$. Dashed lines denote stochastic domain, whereas solid lines denote binary domain. Vector signals are denoted as thick lines.}
    \label{fig_mlp_2_lfsr}
\end{figure}

%---------------------------------------------
% Explicacion de conexion de mlp - 2 LFSR
%---------------------------------------------
Fig.\ref{fig_mlp_2_lfsr} shows how the stochastic FFNN is connected. As noted, only two RNG are used for the whole system, thus saving resources since the RNG circuits are the most area-demanding block in Stochastic Computing design. 
A Linear Feedback Shift Register (LFSR) circuit is used as pseudo-RNG. LFSR1 block generates $R_x(t)$, which is used in the stochastic conversion of the inputs $\mathbf{u}^{1,2}$, the zero reference signal and the APC binary output of each stochastic neuron, thus producing maximum correlation among these signals. 
Binary to Stochastic blocks ($BSC$) are employed to convert from the binary to the stochastic domain using LFSR blocks as reference.
The output vector of each layer is denoted as $\mathbf{a}_{(t)}^{(l)}$, where $l$ denotes the hidden layer of the network.
In order to achieve total decorrelation between neuron inputs
%($\mathbf{u}^{1,2}_{(t)}$ and $\mathbf{a}_{(t)}^{(l)}$) 
and weights, LFSR2 is employed as a second pseudo-RNG (providing signal $R_w(t)$) to generate the stochastic weight vector $\mathbf{w}(t)$. 

As noted, the stochastic hardware implementation presented exploits the correlation phenomenon among signals, and minimizes the area usage by reducing the RNG employed in the circuit to only two LFSR blocks.

%---------------------------------------------
% Figura - mlp 2 LFSR
%---------------------------------------------

%---------------------------------------------
% RESULTS
%---------------------------------------------
\section*{Experiments and Results}
\color{black}
%Up to three levels of \textbf{subheading} are permitted. Subheadings should not be numbered.
%\subsection*{Subsection}
%Example text under a subsection. Bulleted lists may be used where appropriate, e.g.
%\begin{itemize}
%\item First item
%\item Second item
%\end{itemize}
%\subsubsection*{Third-level section}
 %Topical subheadings are allowed.
 
%=======================
% Resultados 
%=======================
% base de datos
In order to evaluate the NN accelerator, we used the DUD-E \cite{dude} database, which contains 22,886 active compounds and their affinities against 102 targets. 
% training and test set
We built the training-set incorporating the 50\% of actives and the 10\% of decoys from each target of the whole data-set. Each data-instance used in the training set incorporates two compounds as inputs (the cristal ligand of the target along with a compound that can be an active or a decoy). Therefore, we finally use a total of 162,530 data-instances in the training-set and 1,300,804 in the test-set.
%The rest of samples are used to build the test-set. 
% Software to Train
%The Neural Network (NN) models were trained using Keras library \cite{chollet2015keras}.
A learning rate of $0.001$ with Adam optimizer was employed for parameter training. 

% Experimentos y comparasiones
We did different comparisons to evaluate our models. 
Firstly, we compared the performance of our software and hardware platforms. 
For software platform models, we used the tangent-hyperbolic function, whereas ReLU function was employed for hardware implementations.
We analyzed ten different FFNN architectures: five for software implementations and five for hardware implementations. 
Finally, the best performance models of each platform were selected to be contrasted with other ligand-based works found in literature. 

% como evaluamos el performance
We evaluated the overall model performance with following metrics: the Area Under the Curve (AUC) of the Receiver Operating Characteristic (ROC) plot, the Enrichment Factor (EF), the processing speed in inferences per second, and the energy efficiency in terms of inferences per Joule. 
The AUC value provides an idea of the overall performance of the model for the full data-set. 
Nevertheless, and considering that VS is a pre-processing step in the drug-discovery process from which only the top-ranked molecules will be finally tested in vitro, other metrics such as the EF are more appropriate. The EF is defined as the ratio between the true-positives rate (TPR) obtained at the top x\% of the ranked database with respect to the expected actives that would be obtained by chance so that $EF=TPR(x\%)\cdot 100/(x\cdot P)$, where $P$ is the total number of positives of a given target in the data-set.
The features used for the compound description are the 12 most significant MPE descriptors explained previously, so that each neural network incorporates an input layer with 24 MPE values as input.
%*********************************************************
% 1. Comparasion mis modelos
%*********************************************************
\subsection*{Software and Hardware models Comparison}

\begin{figure}
    \centering
    \includegraphics[scale=0.3]{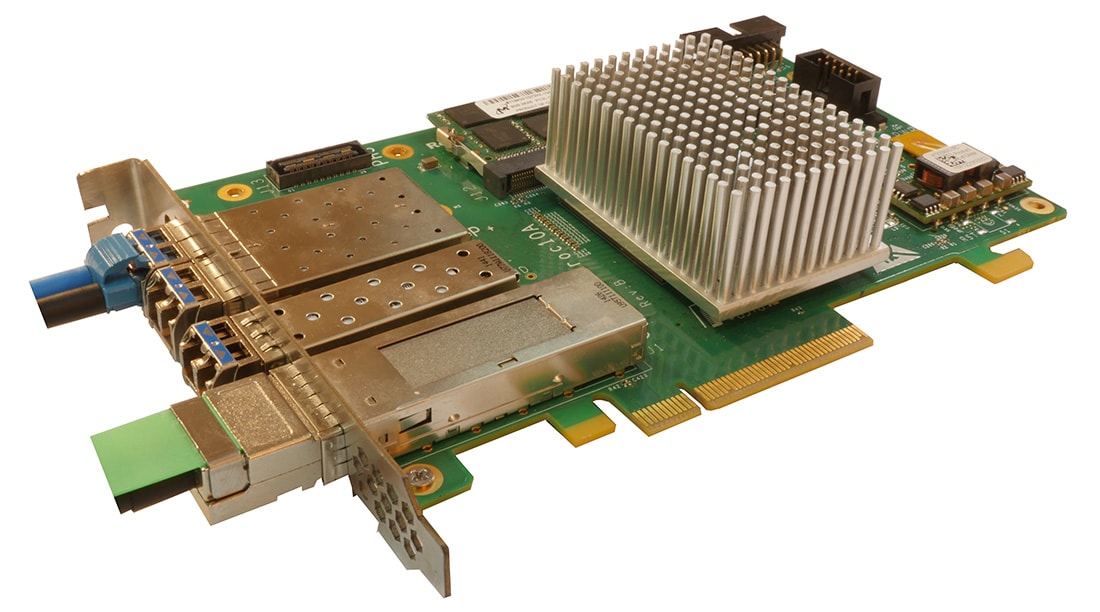}
    \caption{Gidel PROC10A board with an Intel 10AX115H3F34I2SG FPGA\cite{gidel:proc10a_img} running the 12-bit Stochastic Computing (SC) implementation at 125MHz. Board used to evaluate the hardware platform models.}
    \label{fig_gidel_proc10a}
\end{figure}

% explicacion de la tabla.
Table \ref{tab:My_models} shows the performance comparison for both software (Sw) and hardware (Hw) platforms, where the model name indicates the number of neurons in the first hidden layer: 12 for a [12-6-1] , 24 for a [24-12-1], 48 for a [48-24-1], 64 for a [64-32-1] and 256 for a [256-1] network architecture.
% que usamos como hardware y software
The software results were produced using an Intel(R) Xeon(R) X5670 processor with a 64-bit floating-point precision running at 2.93 GHz.
Hardware results were produced using a Gidel PROC10A board (Fig.\ref{fig_gidel_proc10a}), which contains an Intel 10AX115H3F34I2SG FPGA running the 12-bit Stochastic Computing (SC) implementation at 125MHz.
% caracteristicas del hardware
For the hardware implementation, we embedded as many SC model prototypes as could be fitted in the device (reported on the "Parallelization" column).

% analisis de Software Platform
Analyzing the software quadrant, the [64-32-1] model  (Sw 64) presents the best accuracy, with an improvement of $0.07$ AUC compared to the second more accurate (Sw 48). This result is expected considering \textit{Sw 64} is the deeper network from the evaluated models, producing better precision results; however, it is $1.18$x slower than the second more accurate (Sw 48) and $1.38$x slower than the fastest one (Sw 12).
%modelo del 256.
It is interesting to note that although \textit{Sw 256} is the biggest model, it is the poorest in terms of overall performance. 
It presents a degradation of $0.09$ in AUC accuracy and $36\%$ in speed, compared to the more accurate (Sw 64) and faster (Sw 12) models, respectively.

% Analisis de Hardware Platform
Hardware platform quadrant presents lower AUC performance compared to the software platform. 
Different sources contribute to the degradation. 
Firstly, hardware platform presents lower resolution to represent data.
Other difference is the normalization performed at the output of each layer to fit the bit-stream resolution (12 bits in this work). 
Finally, weights might have outliers, that once converted to the \textit{bipolar} stochastic domain [-1,1], may cause a loss of resolution of the rest of weights, thus, producing a degradation with respect to the software training. 
% aun asi muestran mejoras en speed
Even so, hardware models outperform the software ones in terms of speed and energy efficiency, thanks to their parallelizaton and lower power consumption (see Table \ref{tab:My_models}).
% comparasion hardware y software MLP 12
Take as an instance the [12-6-1] architecture, where a degradation of $0.09$ in AUC is observed for the hardware model compared with the software one; whereas an improvement of $10$x is measured in terms of speed and $45$x in energy efficiency.
Similar case is observed when comparing the two highest AUC score models (Sw 64 and Hw 48). 
The hardware model shows a degradation of $0.07$ in AUC performance, whereas an improvement of $2.3$x in speed and $10.4$x in energy efficiency is measured.
An interesting combination process for big data base applications in VS is proposed: the hardware model can be used in the fast front screening phase and the software model in the back, exploiting the advantages from both models.
This could lead to reduce the total time required to find an optimum set of compounds.

%----------------------------
% Tabla 1 MyModels
%----------------------------
\begin{table}[]
    \centering
    \caption{Accelerator performance comparison for software and hardware implementations using different neural network architectures. Higher values per column for each quadrant are noted in bold numbers. AUC column is the mean value calculated among all DUD-E targets. Speed is measured in inferences per second, and energy efficiency in inferences per Joule. Parallelization is accomplished in hardware models fitting the maximum amount of FFNN prototypes that can be implemented in the FPGA.
    }
    \label{tab:My_models}
    \begin{tabular}{lccccc}
    \hline
    model & \multicolumn{1}{l}{AUC} & {\begin{tabular}[c]{@{}c@{}}Speed \\ (inf/sec)\end{tabular}} & {\begin{tabular}[c]{@{}c@{}}Power\\ (W)\end{tabular}} & {\begin{tabular}[c]{@{}c@{}}Energy \\ Efficiency \\ (inf/Joule)\end{tabular}} & {Parallelization} \\ \hline
    
    \multicolumn{1}{l|}{Sw 12} & 0.67 & \textbf{43573} & 95 & \textbf{459} & 1 \\
    
    \multicolumn{1}{l|}{Sw 24} & 0.75 & 42034 & 95 & 442 & 1 \\
    
    \multicolumn{1}{l|}{Sw 48} & 0.78 & 37397 & 95 & 394 & 1 \\
    
    \multicolumn{1}{l|}{Sw 64} & \textbf{0.83} & 31616 & 95 & 333 & 1 \\
    
    \multicolumn{1}{l|}{Sw 256} & 0.74 & 27785 & 95 & 292 & 1 \\
    
    \hline %--------------------------------------------------
    
    \multicolumn{1}{l|}{Hw 12} & 0.58 & \textbf{436364} & 21 & \textbf{20779} & \textbf{72} \\
    
    \multicolumn{1}{l|}{Hw 24} & 0.62 & 163636 & 21 & 7792 & 27 \\
    
    \multicolumn{1}{l|}{Hw 48} & \textbf{0.76} & 72727 & 21 & 3463 & 12 \\
    
    \multicolumn{1}{l|}{Hw 64} & 0.69 & 42424 & 21 & 2020 & 7 \\
    
    \multicolumn{1}{l|}{Hw 256} & 0.70 & 18182 & 21 & 866 & 3 \\ 
    
    \hline
    
    \end{tabular}
\end{table}

% Figura de GIDEL

\begin{figure}
    \centering
    \includegraphics[scale=1.0]{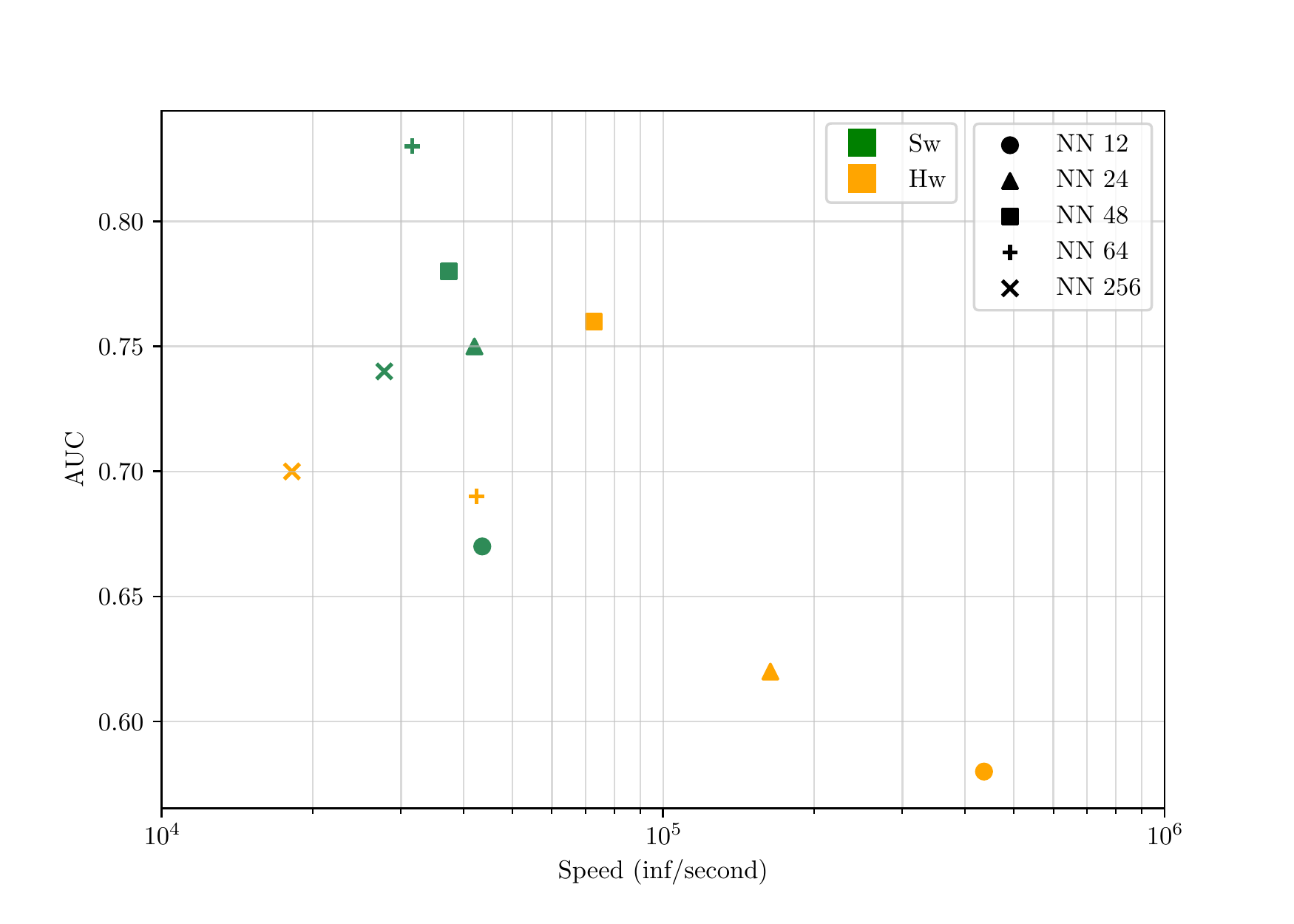}
    \caption{AUC vs inference speed for different software and hardware accelerator models. The speed  for hardware implementations is considerably greater than the software implementations. It is also observed that hardware models show a higher speed improvement when network parameters are modified. }
	\label{fig_auc_vs_variables}
\end{figure}

%----------------------------
% Figura Speed
%----------------------------
Fig.\ref{fig_auc_vs_variables} shows the relationship between AUC performance and speed for both software and hardware models.
As shown, gaps between hardware models in the horizontal axis are longer than those for software ones, showing the higher speed improvements when network parameters are modified.
%The software average speed performance is of $36841$ for the software models and $146667$ for the hardware accelerator. 
The higher speed improvement obtained with the hardware acceleration is achieved at the cost of accuracy degradation.

\begin{figure}
    \centering
    \includegraphics[scale=0.758]{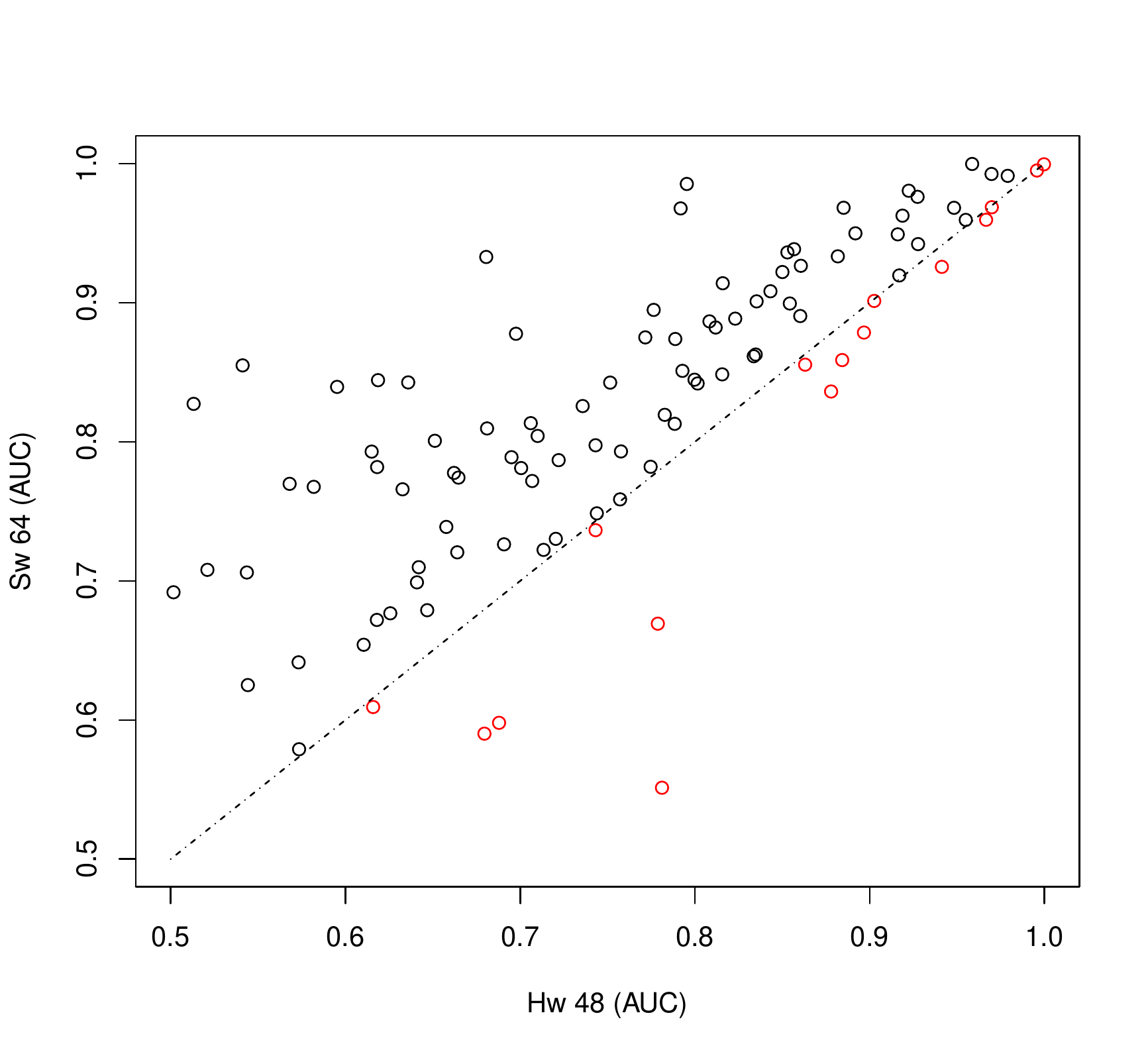}
    \caption{AUC obtained for both models (Sw 64 and HW 48) per each of the 102 DUD-E targets. The Sw 64 model outperforms Hw 48 in most of the studied targets except on 16 cases. Interestingly, for targets where Sw 64 obtains a predictive models of low performance (0.5 < AUC < 0.65), Hw 48 provides substantially better models for three targets (jak2, mk14 and ppard).}
    \label{fig_auc_bars_farmacos}
\end{figure}

%----------------------------
% Figura AUc Bars
%----------------------------
Fig.\ref{fig_auc_bars_farmacos} plots the comparison between the best models of this work in terms of AUC per target. 
For clarity purposes, DUD-E targets are sorted by AUC performance in the software model. As can be observed, AUC performance from hardware model outperforms software model in 28 targets; an interesting detail to take advantage of in case of fast specific compound searching.

%\begin{figure}
%    \centering
%    \includegraphics[scale=0.758]{Images/fig_auc_bars_farmacos.pdf}
%    \caption{AUC for each target in the DUD-E virtual screening benchmark for the two reference models of this work. Targets are sorted by AUC performance in the software model. Hardware model outperforms software model on 28 targets.}
%    \label{fig_auc_bars_farmacos}
%\end{figure}

%*********************************************************
% 2. Comparasion otros papers
%*********************************************************
\subsection*{Comparison with other ligand-based models}

Table \ref{tab:comparison_others} compares the relevant accelerators of this work with four different ligand-based methods from literature : eSim-pscreen\cite{Cleves2019865}, eSim-pfast\cite{Cleves2019865}, eSim-pfastf\cite{Cleves2019865}, and mRAISE\cite{vonBehren2017335}.
We compared AUC, different Enrichment Factors (EF) percentage ($1\%, 5\% $ and $10\%$), as well as two accelerator performance metrics: speed  and energy efficiency when processing the DUD-E dataset. 
% Analisis de AUC
Software model of this work (Sw 64) outperforms all other methods in terms of AUC. 
Comparing to the best AUC method of other works (eSim-pscreen with $0.76$ AUC), it has a $0.07$ improvement. 
Our hardware reference model holds competitive in terms of AUC, performing the same precision than the best eSim method (pscreen)\cite{Cleves2019865}.
% Analisis EF
Software 64 remains as the best method in terms of enrichment factor (EF) except for the 1\%
case, where the mRAISE result is higher by $2.74$.
% Analisis de Speed 
The outstanding difference is observed in terms of speed. Our software model, with  a speed of $31$K inferences per second, performs $115$x faster than the faster eSim work. On the other hand, our hardware model, with a speed of $72$K inferences per second, performs $264$x faster, showing the benefits contributed by employing accelerators for VS.
% analisis energy 
%Energy efficiency is displayed just for the two models of this work considering there are no data for other works.

%-------------------------------------
% Tabla Comparacion otros papers
%-------------------------------------
\begin{table}[]
    \centering
    \caption{Performance comparison between the two relevant models of this work and four different ligand-based methods taken from the literature. Best results per column are noted in bold numbers. Our software model outperforms in terms of precision, while the hardware model exceeds in terms of speed and energy efficiency.}
    \label{tab:comparison_others}
    \begin{tabular}{lccccccc}
    \hline
    Method & \begin{tabular}[c]{@{}c@{}}AUC \end{tabular} & \begin{tabular}[c]{@{}c@{}}EF 1\%\\ \end{tabular} & \begin{tabular}[c]{@{}c@{}}EF 5\%\\\end{tabular} & \begin{tabular}[c]{@{}c@{}}EF 10\%\\\end{tabular} & \begin{tabular}[c]{@{}c@{}}Speed\\ (inf/sec)\end{tabular} & \begin{tabular}[c]{@{}c@{}}Energy\\ efficiency\\ (inf/Joule)\end{tabular} \\ \hline
    
    \multicolumn{1}{l|}{This work (Sw 64)} & \textbf{0.83} & 20.71 & \textbf{9.08} & \textbf{5.63} &  31616 & 333 \\
    
    \multicolumn{1}{l|}{This work (Hw 48)} & 0.76 & 15.07 & 6.69 & 4.42 & \textbf{72727} & \textbf{3463} \\
    
    \multicolumn{1}{l|}{eSim-pscreen\cite{Cleves2019865}} & 0.76 & -- & -- & --  & 12.3 & -- \\
    
    \multicolumn{1}{l|}{eSim-pfast\cite{Cleves2019865}} & 0.74 & -- & -- & --  & 61.2 & -- \\
    
    \multicolumn{1}{l|}{eSim-pfastf\cite{Cleves2019865}} & 0.71 & -- & -- & -- & 274.9 & -- \\
    
    \multicolumn{1}{l|}{mRAISE\cite{vonBehren2017335}} & 0.74 & \textbf{23.45} & 7.78 & 4.69 & -- & -- \\
    
    \hline
    \end{tabular}
\end{table}

%----------------------------
% Tabla threshold
%----------------------------
Table \ref{tab:MyModels_threshold} presents the percentage of targets for the DUD-E benchmark on which different methods fit the given AUC thresholds. In every column, the best performance value is noted in bold.
% Software
The software 64 model produced the best results. It is the unique method with $0\%$ of targets performing worst than random (first column in the table), and is presenting the highest percentage of targets for the rest of AUC threshold values. 
% Hardware
It is interesting to note that the proposed hardware model presents only 1\% of targets performing worst than random, compared to the $5\%$ of targets from the eSim work. For the percentage of targets performing an AUC more than $0.95$ (last column), hardware model performs the same than the best eSim method, thus, producing similar results with the advantage of being $5912$x faster. 

\begin{table}[]
    \centering
    \caption{Comparison of DUD-E targets percentage on which different methods fit the given AUC thresholds. Best values per column are noted in bold numbers. The proposed software model outperforms the rest of the methods. The hardware implementation provides competitive results against the best eSim method, with the advantage of being $264$x faster.}
    \label{tab:MyModels_threshold}
    \begin{tabular}{lcccccc}
    \hline
    
    \begin{tabular}[]{@{}c@{}}Model \end{tabular} & \begin{tabular}[c]{@{}c@{}}\% AUC\\$<$ 0.5 \end{tabular} &
    \begin{tabular}[]{@{}c@{}}\% AUC\\$\geq$ 0.6 \end{tabular} & 
    \begin{tabular}[c]{@{}c@{}}\% AUC\\$\geq$ 0.7 \end{tabular} & 
    \begin{tabular}[c]{@{}c@{}}\% AUC\\$\geq$ 0.8 \end{tabular} & 
    \begin{tabular}[c]{@{}c@{}}\% AUC\\$\geq$ 0.9 \end{tabular} & 
    \begin{tabular}[c]{@{}c@{}}\% AUC\\$\geq$ 0.95 \end{tabular} \\ 
    \hline
    
    \multicolumn{1}{l|} {This work (Sw 64)} & \textbf{0} & \textbf{96} & \textbf{85} & \textbf{61} & \textbf{29} & \textbf{15} \\ 
    
    \multicolumn{1}{l|}{This work (Hw 48)} & 1 & 86 & 63 & 38 & 16 & 8 \\  
    
    \multicolumn{1}{l|}{eSim-pscreen\cite{Cleves2019865}} & 5 & 81 & 69 & 43 & 17 & 8 \\  
    
    \multicolumn{1}{l|}{eSim-pfast\cite{Cleves2019865}} & 9 & 82 & 62 & 34 & 14 & 5 \\  
    
    \multicolumn{1}{l|}{eSim-pfast\cite{Cleves2019865}} & 5 & 79 & 53 & 26 & 6 & 3 \\

    \hline
    \end{tabular}
\end{table}

%*********************************************************
%
% 3. Comparasion DUD38
%
%*********************************************************
Finally, we checked how the Sw-64 model behaves when a specific neural network is trained per each target. 
For comparison purposes, we used the DUD-38 database, which consists of a subset of 38 targets in the DUD-E database that are common to the older DUD database. This comparison is also made by Bonanno and Ebejer \cite{Bonanno20}.
% Distribucion de training
Inspired by them \cite{Bonanno20}, we selected an $80\%$ from the data-set for training and the remaining $20\%$ for testing, applying the oversampling technique to train the imbalanced issue from the data-set. 
% Comparasion en la tabla
After obtaining specific weights per target for the model, we obtained the results shown in Table \ref{tab:DUD}. 
As can be appreciated, the proposed model provides similar accuracy values compared to the one provided by Bonanno and Ebejer (NN-500), taking into account the model used by them is five times bigger, having $500$ neurons in the hidden layer.
% STD comparison
The AUC standard deviation of the proposed model is $6.87$x better than the NN-500 reference, thus, reducing the range of the possible real AUC value of the model. 
Similar case is observed in the EF standard deviation, where an improvement of 2x is obtained.

% TABLE
\begin{table}[]
    \centering
    \caption{Software model Sw-64 and reference \citen{Bonanno20} model comparison for the DUD-38 data-set. Specific training per target evaluation was employed instead of doing a generic training. Reference \citen{Bonanno20} model has 500 neurons in the hidden layer, five times bigger than the model presented in this work.}
    \label{tab:DUD}
    \begin{tabular}{lcc}
    \hline
    {model} & AUC & EF 1\% \\ 
    \hline
    
    \multicolumn{1}{l|}{This work (Sw 64)} & $0.94 \pm 0.048$ & $30.14 \pm 6.95$  \\
    
    \multicolumn{1}{l|}{NN-500 \cite{Bonanno20}} & $0.95 \pm 0.33$ & $37.3 \pm 14.7$  \\
     \hline
    \end{tabular}
\end{table}

%---------------------------------------------
% DISCUSSION AND CONCLUSION
%---------------------------------------------
%\section*{Discussion}
% Usar Hw implementation como una capa de preprocesamiento
\section*{CONCLUSIONS} 
In this work we have shown the powerful combination of the proposed energy-based model with Artificial Neural Network (ANN) to establish similarities between different compounds from the point of view of their expected activities. At the same time we showed how to accelerate the VS process by using FPGAs to improve the performance in terms of speed and energy-efficiency.
As a summary, two different methods are presented in this work, a
software implementation presenting a high overall accuracy and a hardware platform with special high-performance characteristics.
When processing the DUD-E dataset with the same weights (generic training), the software platform presents an AUC of 83\% and an Enrichment Factor of ($EF_{1\%}=20.71$), while hardware presents the best speed and energy-efficiency thanks to the use of an unconventional computing methodology (stochastic computing), with optimal characteristics to accelerate ANNs.
Compared to other previously-publised virtual screening methods, the proposed model improves the AUC value in 0.07, the $EF_{5\%}$ value in a factor of 1.16, and the processing speed in a factor of 115x.
We also presented the energy efficiency for the proposed models: 333 inferences per Joule for the software model, and 3463 inferences per Joule for the hardware model. 
In general the hardware model presents about one order of magnitude better performance than the software implementation, representing a feasible alternative for the processing of huge molecular databases. 
Finally, we evaluated the performance of the proposed energy-based ANN model when considering a specific training for each therapeutic target. 
The AUC value improves from 83\% (for the generic training) to 95\%, while the 1\% enrichment factor increases from 20.71 to 30.14. 

\color{black}

%---------------------------------------------
% BIBLIOGRAPHY
%---------------------------------------------
%\bibliography{sample}
\bibliography{01-main}

%---------------------------------------------
% Acknowledgements
%---------------------------------------------
\section*{Acknowledgements (not compulsory)}

%---------------------------------------------
% AUTHOR contributions
%---------------------------------------------
\section*{Author contributions statement}

J.R. conceived the model and the experiments,  C.F., J.R. and C.B. conducted the experiments, P.B., J.R, V.C. and M.R. analysed the results.  All authors reviewed the manuscript.

\end{document}